\documentclass[11pt]{article} 

\usepackage{url,hyperref,lineno,microtype,subcaption}
\usepackage[onehalfspacing]{setspace}

\usepackage{graphicx,amsthm,amsmath,wrapfig, caption,svg,amssymb,tabularx,bbm,longtable,mathtools,float,cleveref,diagbox,algpseudocode,algorithm,enumitem,color,lipsum,multirow,comment,booktabs,array,bm,authblk,xcolor}

\usepackage[square, comma, sort&compress, numbers]{natbib}
\usepackage[toc,page]{appendix}
\usepackage[font=small,labelfont=bf]{caption}
\usepackage[margin = 1 in]{geometry}

\begin{document}
\onecolumn

\title{Ensemble machine learning approach for screening of coronary heart disease based on echocardiography and risk factors} 

\author{Jingyi Zhang$^{a,1 }$, Huolan Zhu$^{a,2 }$, Yongkai Chen$^{1}$, Chenguang Yang$^{2}$, Huimin Cheng$^{1}$, Yi Li$^{2}$, Wenxuan Zhong$^{\ast,1 }$ and Fang Wang$^{\star,2}$
\thanks{$^{a}$ these authors contributed equally. $^{1}$ Department of Statistics, University of Georgia, Athens, GA 30622 . $^{2}$ Division of Cardiology, Beijing Hospital, Beijing, China. ${^\ast }$ To whom statistical methodology correspondence should be addressed (e-mail: wenxuan@uga.edu). ${^\star }$ To whom clinical methodology correspondence should be addressed (e-mail: bjh\_wangfang@163.com).}}
\doublespace


\date{}
\maketitle

\noindent\textbf{Corresponding authors}: Fang Wang (bjh\_wangfang@163.com) and Wenxuan Zhong (wenxuan@uga.edu)



\begin{abstract}

\textbf{Background:} Extensive clinical evidence suggests that a preventive screening of coronary heart disease (CHD) at an earlier stage can greatly reduce the mortality rate. We use 64 two-dimensional speckle tracking echocardiography (2D-STE) features and seven clinical features to predict whether one has CHD. 

\textbf{Methods:} We develop a machine learning approach that integrates a number of popular classification methods together by model stacking, and generalize the traditional stacking method to a two-step stacking method to improve the diagnostic performance.

\textbf{Results:} By borrowing strengths from multiple classification models through the proposed method, we improve the CHD classification accuracy from around 70\% to 87.7\% on the testing set. The sensitivity of the proposed method is 0.903 and the specificity is 0.843, with an AUC of 0.904, which is significantly higher than those of the individual classification models. 

\textbf{Conclusions:} Our work lays a foundation for the deployment of speckle tracking echocardiography-based screening tools for coronary heart disease.

\end{abstract}

\section{Background}
Coronary heart disease (CHD) is a global epidemic. It led to around $18$ million (roughly one-third of) deaths worldwide in the year 2016 \citep{RN2, RN4,RN1, RN3}. 
Preventive screening of CHD at an earlier stage can significantly reduce the mortality rate, improve the prognosis, and provide therapeutic guidance for patients \citep{RN5}.  Despite urgent needs, an efficient and  effective  screening procedure is still lacking. The majority of CHD diagnostic procedures are radiology-based approaches such as the computed tomography angiography (CTA) and the coronary angiography (CA). These methods can directly visualize the coronary artery and quantify the level of artery occlusion. As a result, these methods are considered the gold standard for diagnosis. Though the radiology-based methods are fairly effective in the CHD diagnosis, their applications in preventive practice are severely limited by the high operational cost, the requirement of expensive and high-maintenance equipment, the need for experienced medical staffs, and  potential side effects\citep{nicholls2019cardiologists}.

A much less explored alternative is the echocardiography-based diagnosis methods, which are commonly used to visualize the movements of the myocardium. In fact, clinical practice suggests that some echocardiology-based techniques, such as the two-dimensional speckle tracking echocardiography (2D-STE) \citep{blessberger2010twod}, can indeed prognosticate CHD. Accumulating evidence shows that some dynamic features extracted by the 2D-STE, such as the global longitudinal strain \citep{RN21} and the time-to-peak strain change, differ significantly between CHD patients and non-CHD patients \citep{RN11}. These observations suggest that the 2D-STE holds a new promise for the CHD screening \citep{blessberger2010two}. However, effective assessment models that can single out early-stage CHD patients with adequate sensitivities and specificities are still lacking. It remains unknown which set of echocardiography-based features can effectively quantify the significance of the myocardial change in response to a minor myocardial anomaly. The requirement of the laboratory-based practice, as opposed to the in-field and real-time analysis, limits their utility for the large-scale population practice. 

The rapid development of machine learning (including computer vision) techniques has triggered a medical technology revolution. For example, the first clinical-grade computational pathology algorithm was proposed in \cite{campanella2019clinical} for the diagnosis of three types of cancers with an average accuracy of 98\%. 
In recent years, machine learning methods were applied to processing images of echocardiograms. These methods, such as convolutional neural networks (CNNs), can help extract image structures and features that are valuable in diagnosis \citep{gulshan2016development, esteva2017dermatologist, litjens2017survey}. For example, CNNs are trained to automatically classify views of echocardiograms, and to extract features from echocardiograms to detect certain diseases \citep{madani2018fast, zhang2018fully}. 
Besides the applications in image segmentation and interpretation, machine learning methods are also expected to play a pivotal role in assisting highly skilled personnel in disease diagnosis by utilizing a series of quantitative, reproducible, and multiplexed features extracted from large amounts of clinical practice. Machine learning methods can capture the potential connection between the features and the diagnosis. For example, in \cite{narula2016machine}, the majority voting method \citep{james1998majority} is applied in distinguishing the hypertrophic cardiomyopathy from physiological hypertrophy in athletes using expert-annotated speckle-tracking echocardiographic features. 

In this article, we aim to develop a machine learning method that takes echocardiographic features as input and classifies whether the subject has CHD. There are many machine learning methods that can be employed to develop a classification method. 
Existing classification methods have various underlying model assumptions, 
which hold the key to the success of the methods. When the data is highly heterogeneous and noisy, as is the case for the echocardiographic data that we analyze, it is not clear which method is suitable as the underlying assumptions are usually hard to validate.  Furthermore, no single classification method provides satisfactory prediction results. 

To improve the classification performance, we integrate 14 classification methods together by an ensemble learning method to provide the best prediction. Through the ensemble learning method, we thus aggregate the strength of all 14 individual classifiers to build the final prediction model. 
In particular, we generalize the traditional stacking method to a two-step stacking method. The first-step stacking can improve the individual prediction by aggregating diversified classifiers; by randomly partitioning the training set multiple times for the second-step stacking, we can reduce the classification errors caused by wrong model aggregation, and weaken the effects of the poor performance of individual classifiers. 

\section{Methods}
In this section, we first present the data used in our study, then briefly review the machine learning applications in echocardiographic analysis and the ensemble learning methods, and finally propose the two-step stacking method.

\subsection{Human subjects}
Our study was a retrospective study based on the clinical trial (NCT03905200). From March 1, 2019 to August 30, 2019, 555 patients were admitted for coronary angiography as suspicious CHD patients. Patients older than 18 were enrolled with written consents. The documentary evidence can be provided if required. 
We excluded patients with non-sinus rhythms, severe heart diseases other than CHD, or other extremely severe organ illnesses.

The echocardiograms were recorded by one experienced clinician on a GE Vivid E9 system (GE Medical Systems, Horten, Norway). Patients’ images were stored in the same machine. Images were transported to an offline EchoPac system of version 201 (GE Healthcare, Horten, Norway), and were further analyzed by an experienced investigator. 
We then excluded patients with low-quality images that EchoPac has troubles in processing.

The study has been performed in accordance with the Declaration of Helsinki, and was approved by the Ethics Committee of the Beijing Hospital.

\subsection{Data and features}
There were $555$ patients examined by a CA or a coronary CTA. Among the $555$ patients, $424$ of them had an echocardiography one day before the angiography was conducted. Patients with vessel stenosis of at least 50\% in the major coronary artery or at least one of its main branches were considered as CHD positive patients \citep{roffi20162015}. Based on such criteria, $217$ of  those $424$  patients are CHD positive. 

For each patient, the recorded  echocardiography consists of three parasternal short-axis standard sections: the mitral valve section, the papillary muscle section, and the apical section, as well as three standard apical sections: the four-chamber view section, the two-chamber view section, and the longitudinal long-axis view section. 
The left ventricular wall (LVW) is divided into 17 segments based on the standard American Heart Association (AHA) 17-segment model \citep{american2002standardized}, each of which has been analyzed individually. Peak systolic longitudinal and radial strains are assessed in all 17 segments to quantify the shortening and thickening of the myocardium for each segment, respectively. The epicardium and endocardium of the left ventricle (LV) are traced automatically and adjusted manually if necessary at the end-systole. The mid-myocardial border is determined at the midpoints between the endocardial and the epicardial borders. The regions of interest (ROIs) cover the endocardium, the myocardium, and the epicardium. The ROIs have been locally adjusted if they are off-track.

In the 2D-STE echocardiography, the most important parameter is the strain, which quantifies the deformation of the myocardium by recording the contractions. 
Since the ventricular contractile dysfunction occurs prior to the electrocardiogram (ECG) change in the sub-endocardium, the diagnostic accuracy based on strains tends to be higher than ECG, troponin, and GRACE score \citep{RN9}.
The longitudinally orientated myocardial fibers are the most susceptible to ischemia \citep{RN21, RN25}. Therefore, the global longitudinal strain has been recommended as the index with the top priority in diagnosing cardiac diseases \citep{RN26, RN27}. 
It is shown in \cite{RN29} that the GLPS can successfully predict CHD (AUC=0.92) for patients with non–ST-segment elevation acute coronary syndromes (NSTE-ACS). 
In the myocardium, micro-vascular communications are network structured. The communication can form some dual arterial perfusion zones. Simply relying on one single index might be inaccurate to decide the etiology. The assessment of myocardium ischemia can be measured by the global longitudinal strain, the global radio strain, the peak systolic strain (PSS), the systolic strain rate (SSR), time to peak (TP), and specific layer strains \citep{RN7, RN28}. 
The myocardium usually consists of three heterogeneous layers of muscle fibers \citep{RN24}. Layer-specific strain is associated with coronary artery disease independently\citep{RN7}. Layer-specific analyses of endocardial, mid-myocardial, and epicardial strains are performed in GLPS as well as the radial strain in the three parasternal short-axis standard sections.

\subsection{Data pre-processing}
 
As shown in table~\ref{features}, we consider 71 features as our predictors for building a machine learning model to predict the risk of CHD, including 64 strain-based numerical features from 2D-STE, age, gender, and five categorical features indicating common risk factors for coronary heart disease. 
According to \cite{torpy2009coronary}, obesity is also a common risk factor for coronary heart disease. However, since the study is a retrospective study, obesity has not been recorded when collecting data. Due to the high correlation between obesity, diabetes, hypertension and, hyperlipidemia \citep{sullivan2008impact}, we include diabetes, hypertension, and hyperlipidemia instead. The other two risk factors we consider are family history and smoking. 
The summary of the clinical characteristics of the subjects is shown in table~\ref{data_summary}, including age, body mass index (BMI), systolic blood pressure (SBP), diastolic blood pressure (DBP), heart rate, gender, hypertension, diabetes, hyperlipidemia, family history, and smoking. From the data summary, we can see that most of the clinical characteristics are balanced between the case group (patients with CHD positive) and the control group (patients with CHD negative). However, we observe a significant increase in the proportion of smoking subjects in the case group when compared with the control group. This observation supports the intuition that smoking is a common risk factor for coronary heart disease.
For the 64 numerical features from 2D-STE, we compare the differences of each feature between the case group and the control group through the two-sample $t$-test \citep{cressie1986use}. The testing results show how significantly CHD can have impacts on each feature. 
To reduce the dimension of features, we apply the principal component analysis (PCA) \citep{wold1987principal} on the 17 segments of PSS, SSR, and TP.
 

\subsection{Machine learning in echocardiographic analysis}
Machine learning methods have been widely applied in fields of echocardiographic analysis \citep{narula2016machine,gandhi2018automation,zhang2018fully,kwon2019deep,chen2019tan,ghorbani2020deep,seetharam2020role,chang2020machine}. 
Recently, most of the applications of the machine learning methods on echocardiogram focus on image segmentation and interpretation \citep{zhang2018fully, chen2019tan, ghorbani2020deep}. 
The methods can learn the shape and size of the region of interest from a labeled training set \citep{carneiro2011segmentation, zhen2015direct, chen2016iterative, pace2018iterative, dangi2018left, tarroni2018comprehensive, dong2018voxelatlasgan, vigneault2018omega}. 
For example, machine learning methods are applied to analyzing the cardiac structures, such as determining global features that can be used to identify standard views of echocardiograms \citep{madani2018fast}, extracting hidden features to detect heart diseases such as hypertrophic cardiomyopathy \citep{zhang2018fully}, identifying certain local structures like pacemaker lead \citep{ghorbani2020deep}, and recognizing the boundaries of ventricle and atrium \citep{chen2019tan, ghorbani2020deep}. Based on the extracted features, \cite{ghorbani2020deep} shows that the machine learning method can identify severely dilated left atrium and left ventricular hypertrophy, estimate right atrium major axis length and left atrial volume, and predict patient age, gender, weight, and height. 
These studies support the hypothesis that machine learning methods can play a promising role in accelerating the image-based diagnostic process. 
The advantage of applying machine learning methods in analyzing medical images lies in the fact that machine learning methods can not only identify features that can be manually recognized, but also extract hidden-layer features that may be difficult to identify \citep{narula2016machine, gandhi2018automation, kwon2019deep}. 
In this paper, we apply machine learning methods on the strain-based local features of the 17 segments as well as the clinical features to link these features to the diagnosis of CHD through the hidden interactions. More specifically, we use machine learning methods to integrate those features through a data-driven diagnostic system built up by classification models and ensemble learning.

\subsection{Ensemble learning and two-step stacking}

When taking echocardiographic features as input to classify whether the patient has CHD, individual classifiers may not provide satisfactory results, as the echocardiographic data is highly heterogeneous and noisy \citep{zhou2012ensemble}. We thus consider multiple classifiers and apply the ensemble learning method to aggregate the strength of all these classifiers to obtain a more precise result \citep{zhou2012ensemble}.
More specifically, we apply the stacking method in this work, since stacking is particularly popular when the signal-to-noise ratio of the data is low \citep{wolpert1992stacked,breiman1996stacked}. The general idea of the stacking is similar to the ``majority voting'' \citep{james1998majority}. To illustrate the stacking method, we thus first look at the majority voting method. Suppose there are $L$ pre-trained classifiers. For one testing data, each classifier gives one classification result $c_l$, for $l = 1,..., L$. When applying majority voting, one can obtain a final classification result $c_f$ as follows,
\begin{equation}\label{eq:MajorityVote}
    c_f = 1\left(\frac{1}{L}\sum_{l=1}^L c_l \geq 0.5\right),
\end{equation}
where $1(\cdot)$ is an indicator function, or a characteristic function, which equals one if the inequality holds and zero otherwise. 

In \eqref{eq:MajorityVote}, the $L$ classifiers have equal weights. One can generalize the majority voting to the weighted voting \citep{kolter2007dynamic}, 
\begin{equation}\label{eq:Wvote}
    c_f = 1\left(\sum_{l=1}^L w_l c_l \geq 0.5\right),
\end{equation}
where $w_l$ is the weight for  classifier $l, l=1,\ldots, L$. Stacking is a generalized weighted voting method. In stacking, the weights $w_1$ through $w_L$ are trained on a validation set through another layer of learning algorithm, with the predictions of the $L$ classifiers on such validation set as the inputs. 
For example, the ``weights'' can be estimated through a linear regression by minimizing the least square errors. Notice that in stacking, the ``weights'' are estimated by learning algorithms that can be rather complex. As a result, the ``weights'' may be negative \citep{gunecs2017stacked}.
In this study, we apply the random forest algorithm \citep{ho1995random} to estimate the stacking weights.

As illustrated in \eqref{eq:MajorityVote} and \eqref{eq:Wvote}, we can see that in ensemble learning methods, the basic idea is to combine a number of classifiers or learners. Some of the individual learners may be just slightly better than random guesses, thus the individual learners are also referred to as ``weak learners''. Through some combination, the predicting power can be improved, then the ensemble is called a ``strong learner''  \citep{hansen1990neural,schapire1990strength}.
In ensemble learning, the fundamental issue is the diversity of the ``weak learners'' \citep{zhou2012ensemble}. It is expected that we will not gain much from the combination if there are not many differences between the weak learners. In other words, the combination of highly correlated weak learners may still result in a weak learner with little improvement. In ensemble learning, the model diversity plays a more important role than the model accuracy of the individual model. As a result, combining individual models with high accuracy, and those with accuracy relatively low always performs better than only combining the accurate ones \citep{zhou2012ensemble}. However, if some individual models are quite poor, they may degrade the performance of the combination. Thus how to balance the model diversity and individual accuracy is quite challenging in ensemble learning  \citep{schapire1990strength, zhou2012ensemble}. In our study, We consider different classes of models vary from traditional parametric model such as logistic regression to the state-of-art learning process such as the neural network. Furthermore, we generalize the classic stacking method to a two-step stacking method to achieve a trade-off between diversity and accuracy. 
Specifically, in the first step, we train individual classifiers $c_{l}^{(k)}, l=1,\ldots, L$ and the weights $w_l^{(k)}, l=1,\ldots, L$ on the $k$th randomly sampled training data. 
In this step, we have classifiers with multiple levels of performance included to expand the model diversity. 
We repeat this process $K$ times, and denote
\begin{equation}\label{step1stack}
    c^*_k = 1\left(\sum_{l=1}^{L}w_l^{(k)}c_l^{(k)} 
    \geq
    0.5 \right), \quad k=1,\ldots, K. 
\end{equation}
In the second step,  we further stack the $K$ classification results $c^*_k, k=1,\ldots, K$ through the weights $w^*_k, k=1,\ldots, K$ trained on the validation data. 
The second step then can weaken the effects of the poor performance of individual classifiers and reduce the classification errors caused by wrong model aggregation in the first step. 
We then get the final classifier, 
\begin{equation}\label{step2stack}
    c_{stacking} = 1\left(\sum_{k=1}^{K}w^*_kc^*_k \geq
    0.5
    \right).
\end{equation}

\begin{figure}[h]
    \begin{center}
    \includegraphics[scale = .5]{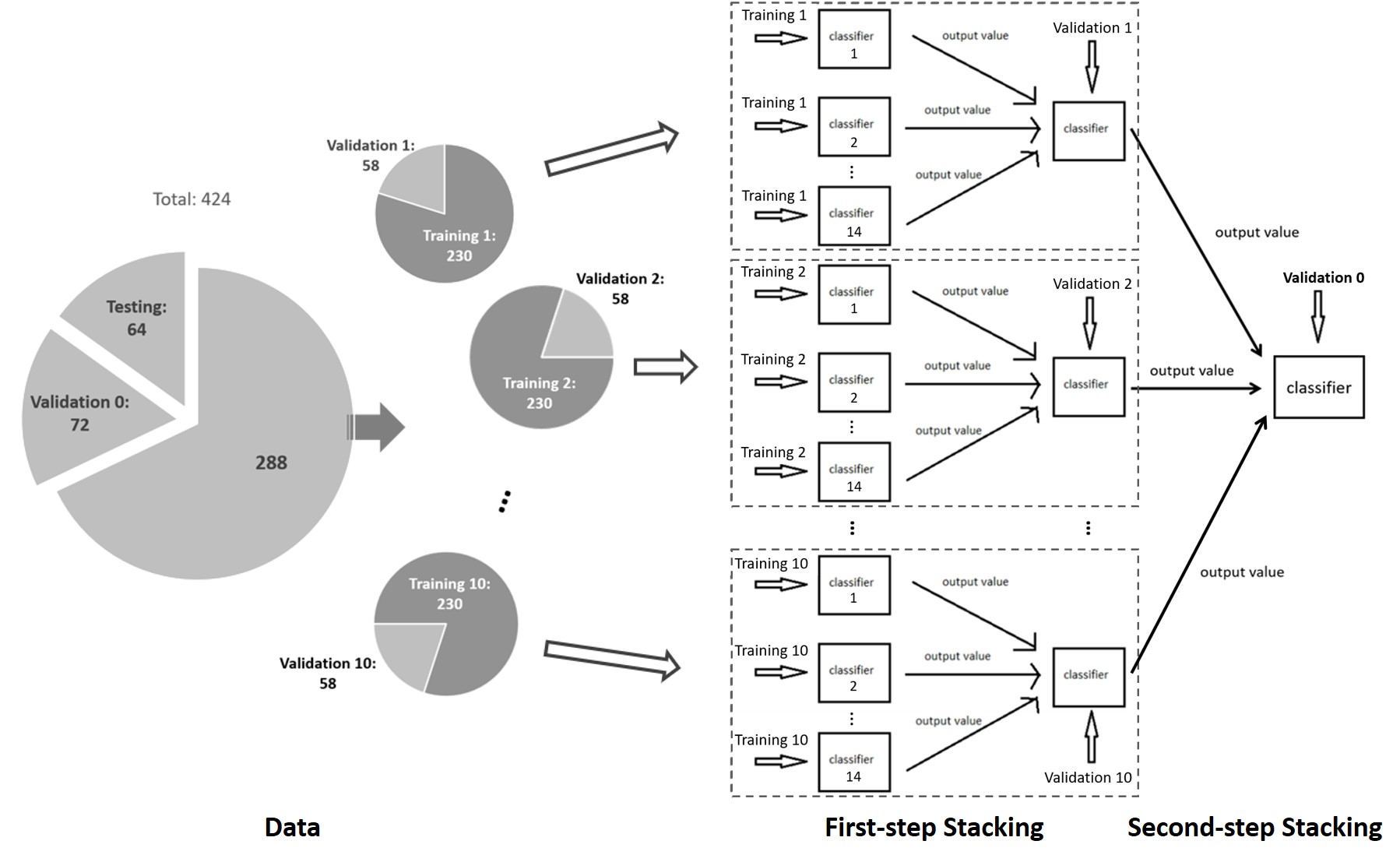}\\
        \caption{Flowchart of the two-step stacking method. The testing set of size 64, named ``Testing'', is used to evaluate the proposed method. The validation set of size 72, named ``Validation 0'', is used to train the second-step stacking weights $w^*_k, k = 1, ...., 10$ in Eq. \eqref{step2stack}. The rest set of size 288 is randomly divided into a first-step training set (named ``Training 1'' through ``Training 10'') of size 230 and a first-step validation set (named ``Validation 1'' through ``Validation 10'') of size 58 to train the 14 individual classifiers $c^{(k)}_l, l = 1, ..., 14$ and first-step stacking weights $w^{(k)}_l, l = 1, ..., 14$ in Eq. \eqref{step1stack} for 10 times. }\label{modelstacking}
    \end{center}
  \vspace{-10pt}
\end{figure}

In particular, as shown in Fig. \ref{modelstacking}, we set $15\%$ of the $424$ subjects as the testing set. Among the remaining $85\%$ subjects, we then set $20\%$ as the validation set and the remaining as the training set for the second-step stacking. For the first-step stacking, we also set $20\%$ of the subjects as the validation set. More specifically, 
we divide the $424$ subjects into a testing set that contains $64$ subjects, a training set that contains $288$ subjects, and a validation set that contains $72$ subjects. For the first step stacking, we repeatedly sample $230$ individuals randomly from the training set as the first-step training set to train the  classifiers $c_l^{(k)}$s in Eq. \eqref{step1stack}, and use the rest of $58$ subjects as the first-step validation set to train the stacking weights $w_l^{(k)}$s in Eq. \eqref{step1stack}. 
In this paper, we build $14$ classifiers using $14$ machine learning approaches, i.e. $L = 14$. We repeat the process $10$ times, i.e., $K = 10$, so that we obtain $10$ classifiers for the second step stacking. The second-step stacking weights $w^*_k$s in Eq. \eqref{step2stack} are trained on the pre-determined validation set of size $72$. To avoid the effects brought by the imbalance of labels through random splitting, we apply the stratify splitting to split the dataset based on the labels so that in each sub-sample, the CHD negative-to-positive ratio remains similar.


\section{Results}

\subsection{Two-sample $t$-test on features}

We compare the differences of GLPS's between the case group and the control group in three layers of the myocardium using a two-sample $t$-test. We record the p-values for the testing. Note that a small p-value indicates a significant difference. 
In this study, we use the threshold p-value $\le 0.05$ to determine if the difference is significant. 
Intuitively, we claim that the CHD has a greater effect on a feature if the difference of such feature between the case group and the control group is more significant. The p-values for the two-sample $t$-test on GLPS's are shown in table~\ref{features_pvalue}. The results confirm that CHD has significant effects on GLPS values. We also conduct the two-sample $t$-test on PSS, SSR, and TP.
From the testing results, we can see that PSS, SSR, and TP are also important features for CHD prediction. When considering the radial strains, the two-sample test results for the radial strains in the apical section (SAX-AP), the papillary muscle section (SAX-PM), and the mitral valve section (SAX-MV) indicate that the radial strain contributes less than the longitudinal strain in CHD prediction (the p-values are all listed in table~\ref{features_pvalue}). 



\subsection{Principal component analysis}

\begin{figure}[h]
    \begin{center}
    \includegraphics[scale = .4]{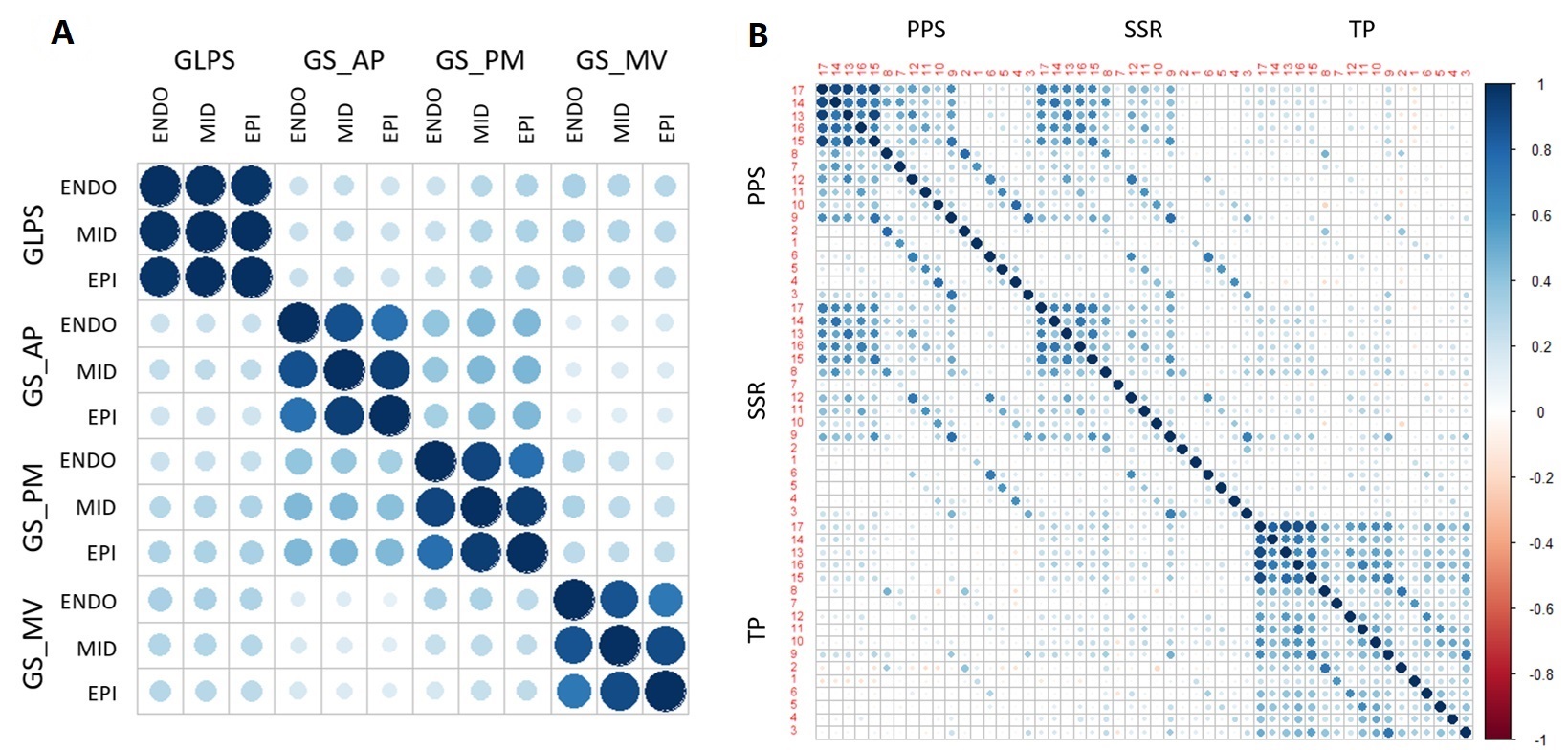}\\
        \caption{Correlations among features. \textbf{(A)}: Correlation matrix of global longitudinal strains and radial strains of apical level, papillary muscle level and mitral valve level. \textbf{(B)}: Correlation matrix of 17 segments on PSS, SSR and TP. 
        }\label{cormat}
    \end{center}
\end{figure}

We first study the correlations among the numerical features. Panel \textbf{(A)} in Fig.~\ref{cormat} shows the correlations between global longitudinal strains and radial strains. We can see that longitudinal strains are weakly correlated with radial strains. For radial strains, each section is weakly correlated with each other. Panel \textbf{(B)} in Fig.~\ref{cormat} shows the correlations among 17 segments on PSS, SSR and TP. From the correlation matrix, we can see that PSS is correlated with SSR, while TP is weakly correlated with both PSS and SSR. When examining the correlation among the 17 segments for PSS, SSR, and TP, respectively, we divide the $17$ segments into apex, apical, mid-cavity, and basal levels based on the AHA 17-segment model, as shown in panel (\textbf{B}) of Fig.~\ref{pcs}. We can see that (i) the apex and apical levels are highly correlated; (ii) for PSS, six segments in the mid-cavity level are highly correlated with their neighboring segments in the basal level; (iii) for SSR, mid-cavity level and basal level are weakly correlated; and (iv) for TP, the correlations among all 17 segments are higher than those in PSS and SSR. Based on the results of the correlation study, we choose to conduct PCA on PSS, SSR, and TP, respectively. 


\begin{figure}[h]
    \begin{center}
    \includegraphics[scale = .3]{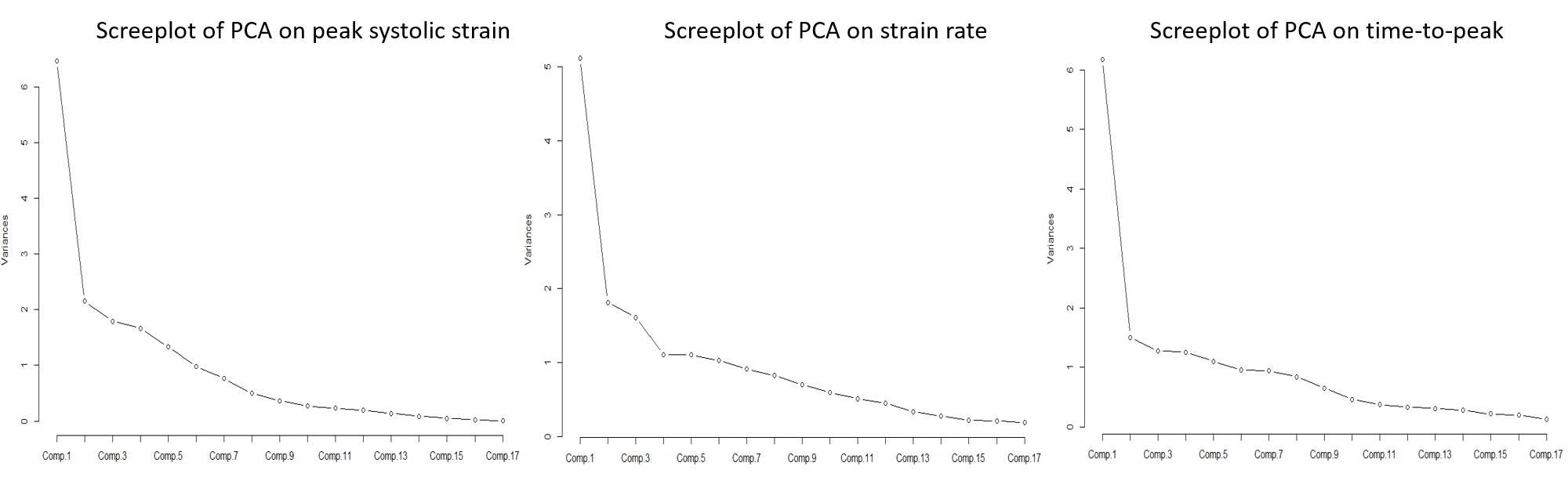}\\
        \caption{Screeplot of PCA on peak systolic strain, systolic strain rate and time-to-peak.}\label{screeplot}
    \end{center}
\end{figure}

\begin{figure}[h]
    \begin{center}
    \includegraphics[scale = .45]{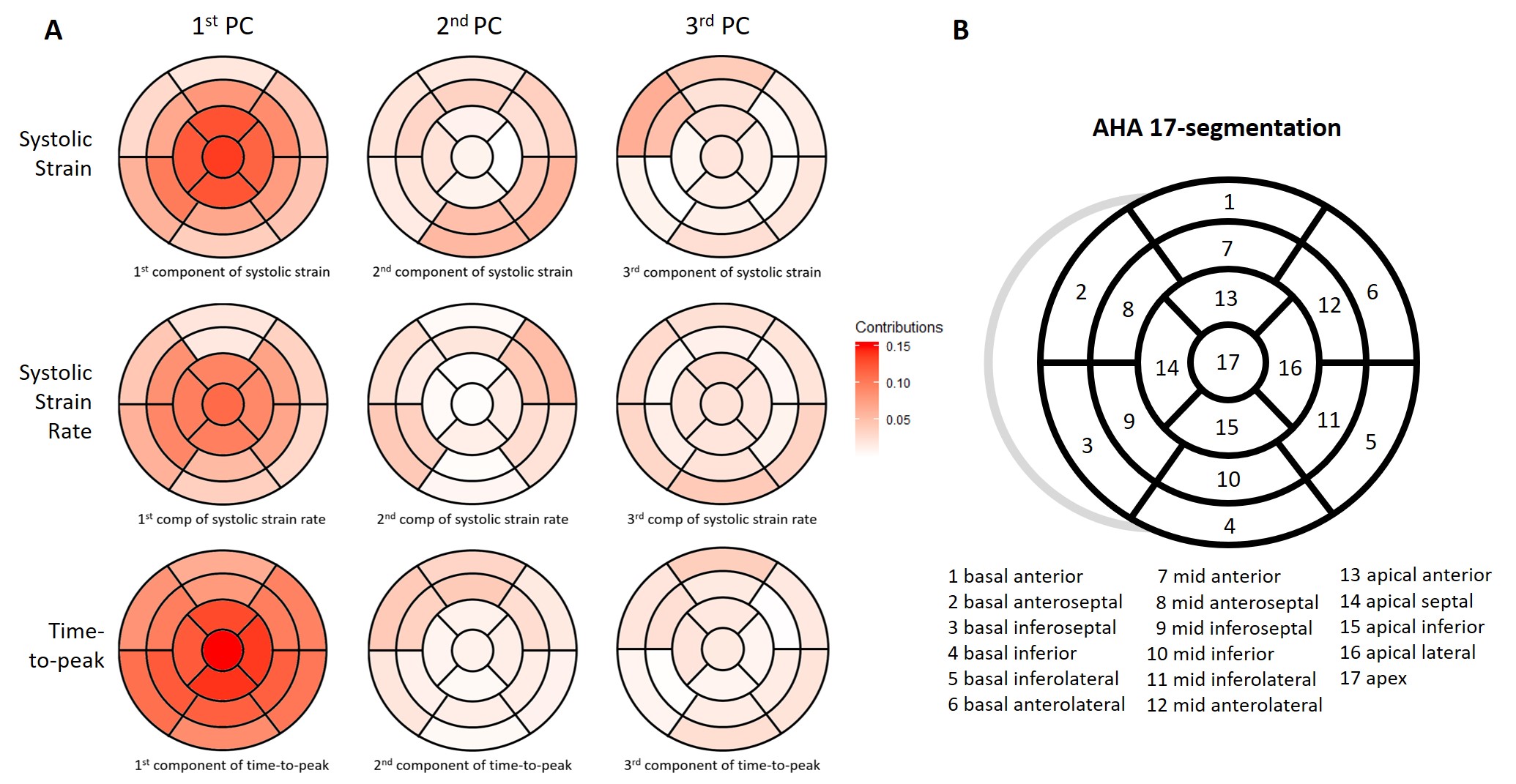}\\
        \caption{(\textbf{A}): Heatmaps of contributions of 17 segments in first three PCs of peak systolic strain, systolic strain rate and time-to-peak. Column from left to right represents the first PC to the third PC respectively, and the top row represents PSS, the middle row represents SSR and the bottom row represents TP. (\textbf{B}): Bullseye plot of the AHA 17-segment model.}\label{pcs}
    \end{center}
\end{figure}

Figure~\ref{screeplot} shows the scree-plots of PCs for features in PSS, SSR, and TP. In each plot, we can find obvious ``elbows", based on which we choose the proper number of PCs to retain in the model. 
Figure~\ref{pcs} shows the heatmaps of the first 3 PC loadings for PSS, SSR, and TP, respectively. From Fig.~\ref{pcs}, we can see that (1) for PSS, SSR, and TP, the first PCs roughly represent the overall average of the $17$ segments. (2) For PSS, the second PC represents the basal/mid inferoseptal, the basal/mid inferior, and the basal/mid inferolateral; the third PC represents the basal/mid anterior and the basal/mid anterolateral. (3) For SSR, the second PC represents the basal/mid anteroseptal and the basal/mid inferolateral; the third PC represents the basal layer. (4) For TP, the second PC represents the basal/mid anterior, the basal/mid anterolateral, and the basal/mid inferolateral; the third PC is similar to the second PC. Thus we choose the first three PCs for PSS and SSR, and the first two PCs for TP.


\subsection{Two-step stacking}

\begin{figure}[h]
    \begin{center}
    \includegraphics[scale = .4]{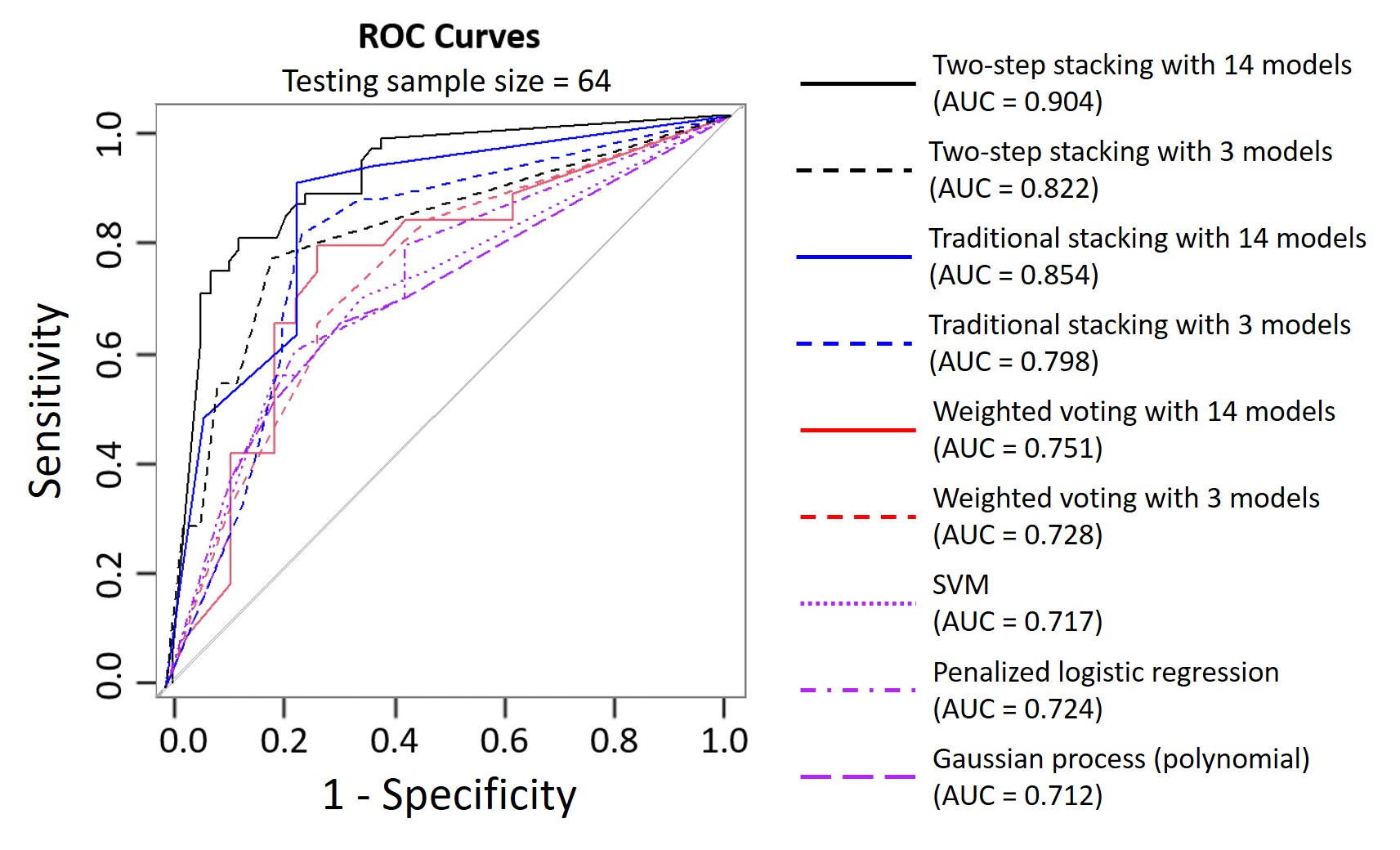}\\
        \caption{ROC curves of 1. the ensemble learning methods on 14 individual models, 2. the ensemble learning methods on the three ``best-perform'' models, and 3. the three ``best-perform'' individual models. The ensemble learning methods including the two-step stacking methods, the traditional stacking methods, and the weighted voting methods. The purple lines represent the individual models. The black lines represent the two-step stacking methods, the blue lines represent the traditional stacking methods, and the red lines represent the weighted voting methods, with the solid lines represents ensemble on 14 models, and the dashed lines represent ensemble on 3 models.}\label{roc}
    \end{center}
\end{figure}

We use the R-package \textit{caret} to build 19 commonly used classifiers. The hyper-parameters for the individual classification model are automatically tuned based on the cross-validation method. 
After 50 replicates, table~\ref{single_results} reports the mean accuracy of all individual classifiers on the testing set, with the standard deviation listed in the brackets. 
We can see that the highest accuracy is $71\%$.
Based on the individual accuracy, we first exclude the five classifiers with the accuracy below $60\%$. For the remaining 14 classifiers, we conduct the ensemble learning method to improve the classification accuracy. Since there is no significant difference among the performance of the remaining 14 models, the question then is how to balance ``model accuracy'' and ``model diversity'' in ensemble learning? To answer this question, we consider the traditional weighted voting method, traditional model stacking, and the proposed two-step stacking on three best-performing individual models with accuracies above $70\%$, and compare the results with those on all the 14 remaining models. The results of 50 replicates are shown in table~\ref{ensemble_results}, with figure~\ref{roc} showing the ROC curves. In Fig.~\ref{roc}, the purple lines present each individual model, the red lines represent the traditional weighted voting method, the blue lines represent the traditional stacking model, and the black lines represent our two-step stacking model. For the three ensemble learning methods, the solid lines represent the ensemble on all 14 models, and the dashed lines represent the ensemble on the three ``best-performing'' models. We then interpret the results from the following three aspects. 
\begin{enumerate}
    \item The stacking methods outperform the weighted voting methods. Such an observation indicates that the stacking method can combine the individual results in a more efficient way.
    \item The 3-model weighted voting only slightly improves the accuracy compared with the individual models. It indicates that the three models may be highly correlated, i.e., the diversity is not enough for a considerable improvement for the ensemble. The 14-model ensemble methods result in a better performance than the 3-model ensemble methods. The results confirm the importance of model diversity in ensemble learning, especially when models are combined through a more complex way in model stacking. 
    \item The traditional model stacking improves the classification accuracy from the $67.3\%$ (the average accuracy for the individual models) to $72.5\%$. Through the proposed two-step stacking, we further improve the classification accuracy to an average of $87.7\%$ on the testing set, with a sensitivity of 0.903 and a specificity of 0.843. In fact, the two-step stacking method significantly outperforms all the other methods.
\end{enumerate}
Based on \cite{RN29}, using GLPS  can successfully predict CHD for NSTE-ACS patients with an AUC of 0.92. We apply our method on GLPS only to see if the accuracy remains. The results are also listed in table~\ref{ensemble_results}, we can see that the accuracy based on GLPS only drops to $63.3\%$ with an AUC of 0.67. Such a drop may be caused by the quality of images in the retrospective study. During the retrospective study, the data were collected during real-time medical treatment, where the priority is efficiency. Thus the data quality may become hard to control. In summary, our method shows the best diagnostic performance in identifying CHD patients among all the methods we compared. The codes for the final 14-classifier two-step stacking model prediction are available in the supplementary materials (additional file 1). 



\section{Discussion} 

\subsection{Clinical implication}
Imaging techniques have been applied  to prognosis and prevention to reduce morbidity and mortality \citep{RN23}. Among all the imaging techniques, echocardiography is one of the most promising techniques in the cardiovascular field. It is noninvasive, convenient, safe, and effective. 2D-STE as a novel technique has its advantage compared with the conventional echocardiography and other modalities. The sub-endocardial myocardial fibers are oriented longitudinally, so the longitudinal myocardial function is affected primarily when ischemia is onset. The decrease in global longitudinal strain, which suggests the ventricular contractile dysfunction, occurs prior to ECG change. Therefore, the machine learning model based on features with the global longitudinal strain included is more efficient than the ECG. Traditionally, the conventional echocardiographic parameters  are mostly estimated by a visual assessment of the ventricular wall contraction in CHD patients. However, subtle abnormalities might be overlooked by human eyes \citep{RN9}. This clinical practice renders  the conventional echocardiography ineffective in the diagnosis of CHD in general and the early stage CHD in particular. Thus, the effectiveness of conventional echocardiography is limited in CHD diagnosis, especially in the early stage. Since the 2D-STE image can detect the tiny abnormalities of the systolic function \citep{RN29,RN30}, it is more promising in CHD diagnosis than the conventional echocardiogram. 

Compared to coronary angiography, our echocardiography-based method can be applied to almost all patients.  Coronary angiography is the gold standard in the diagnosis of stenosis. However, due to its potential medical risks, angiography is not recommended to all patients, such as elder patients, or patients with other end-stage organ failures. 2D-STE helps rule out patients without coronary heart disease and avoid unnecessary coronary angiography. Compared with the time-consuming tests such as MRI and SPECT, our method can provide the diagnosis result in less time. 

The potential clinical applications of the echocardiography-based machine learning method are extensive. Clinicians are always searching for a safer and more effective method for the diagnosis and prognosis of CHD. Studies have shown that the early-stage medical intervention can reduce the mortality and morbidity for CHD \citep{RN31}. 
We believe that our method holds a promise to provide a more efficient and noninvasive early screening and diagnosis of CHD, and could bring a revolutionary impact on the diagnosis modality. 
Moreover, our method based on 2D-STE can also help in re-evaluating the recovery from ischemia after the first hospitalization. It can be recommended as a routine in the physical examination.

\subsection{Method innovation}

Our method is an ensemble learning method.
The ensemble learning methods can be divided into three classes: bagging, boosting, and stacking \citep{zhou2012ensemble}. 
In particular, bagging aims to reduce variance, boosting decreases bias, and stacking improves the prediction. 
Since the goal of this study is to improve the prediction power,
we use the stacking method to aggregate the strengths of popular machine learning methods  \citep{wolpert1992stacked,breiman1996stacked}. 
We generalize the traditional stacking method to a two-step stacking method to achieve a trade-off between the model diversity and accuracy in ensemble learning. 
The first-step stacking aggregates diversified classifiers to improve the individual prediction; the second-step stacking combines multiple first-step stacking classifiers under randomly partitioned training sets to weaken the effects of the wrong model aggregation and the poor performance of individual classifiers. 

\subsection{Limitations}
Our study is a single-center study. The data are collected from the same medical system. Different echo-cardiographic inter-vendors and post-processing algorithms were not applied. The single data-collecting system and the relatively small dataset may increase the instability of the models and lead to low generalizability of the results. 
We have reached an agreement with other hospitals to collect more data from  multiple medical centers. There are potential difficulties in analyzing multi-center data, such as the concerns on the data privacy and data heterogeneity. To overcome these two major concerns, we consider applying the decentralized system. Furthermore, with the multi-center data, we can extend the method to an adaptive learning process so that the model can automatically update when bringing in new samples.
Another limitation is that the speckle tracking analysis can not be conducted automatically. The subjective effects of different physicians might also affect the final prediction. In addition, when processing low-quality images, EchoPac can not recognize the epicardial or endocardial border. Therefore, it may bring certain biases to the results. We are now developing an automatic image quality-control and tracing technique for analyzing echocardiograms. By reducing the user intervention in both image feature extracting and classification analysis, we can effectively minimize subjective errors. 

\subsection{Future works}
With the advantages of machine learning methods in accelerating the image-based diagnostic process, we explore the potential use of machine learning in echocardiographic analysis in the following two aspects.
\begin{enumerate}
    \item \textit{Image quality control}. The machine learning methods are promising in identifying standard views of echocardiograms \citep{madani2018fast, zhang2018fully, chen2019tan, ghorbani2020deep}. When combining with the statistical hypothesis test, we can apply the machine learning methods in echocardiographic quality control. Specifically, the testing method, especially the non-parametric test, can quantify the differences between individual echocardiograms and the ``standard'' echocardiograms utilizing the features extracted by machine learning methods \citep{zhang2018smoothing, xing2020minimax}. Based on the quantified differences, the quality control method can weed out the low-quality images automatically, thus can improve the accuracy in the image-based diagnosis.
    \item \textit{Image segmentation and tracing.} Existing image segmentation methods require a large quantity of annotated training datasets \citep{chen2019tan}. Labeling images, especially medical images, is super labor-intensive and time-consuming. The application of optimal transport, deformation mapping, and transfer learning can help develop a reference-based image segmentation and tracing method. Such a method can detect certain local structures in echocardiograms through a ``transfer'' from the typical annotated references \citep{zhang2020review}. The volume  of the training set thus can be reduced to a size that can be processed in practice.
\end{enumerate}

\section{Conclusion}
Our method enjoys the following practical advantages in screening  CHD. First, our method shows a  good diagnostic performance in identifying CHD patients, i.e., 87.7\% (accuracy), 90.3\% (sensitivity), 84.3\% (specificity). Second, compared with some conventional  CHD diagnosis technologies,  e.g.,  coronary angiography, our method is noninvasive. Our predictive model only requires the 2D-STE features and some commonly used clinical features. 
Third, compared with traditional time-consuming tests, e.g., MRI and SPECT, our method can provide  diagnosis results in significantly less time. 
In summary, our method holds a promise to provide a more efficient and noninvasive early screening and diagnosis of CHD.

\section*{List of abbreviations}

\noindent 
2D-STE: two-dimensional speckle tracking echocardiography\\
AHA: American Heart Association\\
CA: coronary angiography\\
CHD: coronary heart disease\\
CTA: computed tomography angiography\\
ECG: electrocardiogram\\
GLPS: global longitudinal peak strain\\
LV: left ventricle\\
LVW: left ventricular wall\\
PCA: principal component analysis\\
PSS: peak systolic strain\\
ROI: region of interest\\
SSR: systolic strain rate\\
TP: time to peak

\section*{Declarations}
\subsection*{Ethical approval and consent to participate}
All procedures performed in studies involving human participants were in accordance with the ethical standards of the institutional and/or national research committee and with the 1964 Helsinki declaration and its later amendments or comparable ethical standards. Written informed consent was obtained from all individual participants included in the study. The study was approved by Beijing Hospital Ethics Committee (1100000185432).

\subsection*{Consent for publication}
Not applicable.

\subsection*{Availability of data and materials}
One testing dataset supporting the conclusions of this article is included in the supplementary materials (additional file 2). Other datasets and the trained stacking models used and analysed during the current study are available from the corresponding author on reasonable request.

\subsection*{Competing interests}
The authors declare that they have no conflict of interest.

\subsection*{Funding}
Fang Wang is supported by Beijing Municipal Science and Technology Commission for Scientific Research (Z161100000516053). The grant supports the study on the value of speckle tracking technique in the diagnosis and follow-up of coronary heart disease. Fang Wang is also supported by grants from Capital Health Development Research Project (BH2016-071) and the 13th Five-year National Science and Technology Major Project (2017ZX09304026). The fundings had no role in the design of the study, data collection, analysis, or writing of the manuscript.

\subsection*{Authors' contributions}
JZ and HC proposed the two-step stacking method and constructed the predictive model. JZ and YC wrote the code and analyzed the results. 
CY recruited the patients in the clinical trial and the designed experiment. YL summarized the echocardiographic features and reviewed the angiography data. HZ managed and completed the experiment. JZ contributed in the method and analysis parts of the manuscript, and HZ contributed in the clinical part.
FW and WZ conceived the project. All authors contributed to the preparation of the manuscript.

\subsection*{Acknowledgments}
Not applicable.



\bibliographystyle{abbrv}

\bibliography{reference}

\section*{Figure Legends}

\noindent 
Figure 1: Flowchart of the two-step stacking method. \\
Figure 2: Correlations among features. \\
Figure 3: Screeplot of PCA on peak systolic strain, systolic strain rate and time-to-peak.\\
Figure 4: Heatmaps of contributions of 17 segments in first three PCs of PSS, SSR and TP, and the bullseye plot of the AHA 17-segment model.\\
Figure 5: ROC curves of the two-level stacking model, the traditional stacking model, the weighted voting, and individual models. 

\section*{Additional Files}

\noindent 
Additional file 1: stackingModelCode.R; Codes for the final 14-classifier two-step stacking model prediction.\\
Additional file 2: TestingDATA.csv; One testing dataset of size 64. 

\begin{table}[h]
 \caption{Features chosen to be predictors in CHD prediction model.}\label{features}
\resizebox{\textwidth}{!}{
	\begin{tabular}{ lll }
		\hline
		\multicolumn{3}{c}{\textbf{2D-STE features}}\\
		\hline
         & Peak systolic strain (PSS) & 17 segments \rule{0pt}{2.6ex}\\
		Longitudinal strain & Rate of systolic strain (SSR) & 17 segments\\
		 (mid-layer) & Time-to-peak (TP) & 17 segments\\
		\hline 
		 & Mitral valve level (MV) & 3 layers (ENDO/MID/EPI) \rule{0pt}{2.6ex}\\
		Global strain (GS) & Papillary muscle level (PM) & 3 layers\\
		for radio & Apical level (AP) & 3 layers\\
		\hline 
		\multicolumn{2}{l}{Global longitudinal peak strain (GLPS)} & 3 layers (ENDO/MID/EPI) \rule{0pt}{2.6ex}\\
		\hline
        \multicolumn{3}{l}{Peak standard deviation (PSD)} \rule{0pt}{2.6ex}\\
		\hline
		\hline
		\multicolumn{3}{c}{\textbf{Clinic features}}\\
		\hline
        \multicolumn{3}{l}{Age (integer)}\\
		\multicolumn{3}{l}{Gender (M/F)}\\
		\multicolumn{3}{l}{Hypertension (Y/N)}\\
		\multicolumn{3}{l}{Diabetes (Y/N)}\\
		\multicolumn{3}{l}{Hyperlipemia (Y/N)}\\
		\multicolumn{3}{l}{Smoke (Y/N)}\\
		\multicolumn{3}{l}{Family history (Y/N)}\\
		\hline
	\end{tabular}}
\vspace{-10pt}
\end{table}

\begin{table}[h]
 \caption{Summary of clinical characteristics of the subjects.}\label{data_summary}
\begin{center}
\resizebox{0.8\textwidth}{!}{
	\begin{tabular}{ l|ll }
		\hline
         & CHD positive ($n = 217$) & CHD negative ($n = 207$) \rule{0pt}{2.6ex}\\
        \hline
        Age(years) & $64.39 \pm 9.79$ & $64.11 \pm 9.52$\\
		BMI ($\mathrm{kg}/\mathrm{m}^2$) & $25.74\pm3.36$ & $25.56\pm3.75$\\
		DBP & $77.75\pm11.02$ & $79.52\pm11.78$\\
		SBP & $135.83\pm17.56$ & $134.89\pm16.49$\\
		Heart Rate & $75.09\pm11.03$ & $75.70\pm13.32$\\
		Male & $76.50\%$ & $74.32\%$\\
		Hypertension & $66.2\%$ & $67.8\%$\\
		Diabetes & $30\%$ & $41.74\%$\\
		Hyperlipemia & $72.6\%$ & $68.6\%$\\
		Smoke & $52.5\%$ & $28\%$\\
		Family history & $36.1\%$ & $32.5\%$\\
		\hline
	\end{tabular}}
\end{center}
\end{table}

\begin{table}[h]

 \caption{P-values for the two-sample $t$-test of 2D-STE features.}\label{features_pvalue}
\begin{center}
\resizebox{\textwidth}{!}{
	\begin{tabular}{ llllllllll }
		\hline
		\multicolumn{10}{c}{Longitudinal Strain}\\
		 & \multicolumn{3}{c}{GLPS (\textbf{p-value: 0.002})} & PSS & SSR & TP & PSD & & \\
		 \hline
		 & Epi & Mid & Endo & & & & & & \rule{0pt}{2.6ex}\\
		 p-value & .024 & .049 & .076 & .024 & .041 & .179 & .731 & & \\
		 \hline
		 \hline
		 \multicolumn{10}{c}{Radial Strain}\\
         & \multicolumn{3}{c}{ SAX-AP (\textbf{p-value: 0.876})} & \multicolumn{3}{c}{ SAX-PM (\textbf{p-value: 0.503})} & \multicolumn{3}{c}{ SAX-MV (\textbf{p-value: 0.277})}\\
         \hline
         & Epi & Mid & Endo & Epi & Mid & Endo & Epi & Mid & Endo \rule{0pt}{2.6ex}\\
         p-value & .982 & .952 & .598 & .663 & .654 & .682 & .247 & .175 & .516\\
		\hline
	\end{tabular}}
\end{center}
\vspace{-10pt}
\end{table}

\begin{table}[h]
 \caption{Mean testing accuracy of individual classification models after 50 replicates with standard deviation in the brackets.}\label{single_results}
\begin{center}
\resizebox{0.8\textwidth}{!}{
	\begin{tabular}{ ll }
		\hline
         Model & Accuracy \rule{0pt}{2.6ex}\\
        \hline
		logistic regression & $67.7\% (0.034)$\\
		penalized logistic regression & $70.8\% (0.022)$\\
		cumulative probability model & $68.6\% (0.035)$\\
		random forest & $59.2\% (0.034)$\\
		weighted subspace random forest & $59.3\% (0.033)$\\
		SVM with class weight & $70.2\% (0.043)$\\
		SVM with polynomial kernel & $66.3\% (0.041)$\\
		SVM with radial kernel & $63.7\% (0.041)$\\
		K-nearest neighbor & $58.2\% (0.037)$\\
		LDA & $69.6\% (0.048)$\\
		sparsed LDA & $58.8\% (0.036)$\\
		naive Bayes & $64.4\% (0.024)$\\
		Bayes generalized linear model & $68.0\% (0.031)$\\
		Gaussian process with polynomial kernel& $70.1\% (0.035)$\\
		Gaussian process with radial kernel & $65.2\% (0.029)$\\
		Neural network & $62.8\% (0.043)$\\
		Monotone multi-layer perceptron neural network & $69.2\% (0.026)$\\
		model average neural network & $65.1\% (0.035)$\\
		stochastic gradient boosting & $57.8\% (0.027)$\\
		\hline
	\end{tabular}}
\end{center}
\end{table}

\begin{table}[h]
 \caption{Mean testing accuracy and the AUC of ensemble learning methods after 50 replicates with standard deviation in the brackets.}\label{ensemble_results}
\begin{center}
\resizebox{0.85\textwidth}{!}{
	\begin{tabular}{ lll }
		\hline
         Model & Accuracy & AUC \rule{0pt}{2.6ex}\\
        \hline
		\textbf{Two-step stacking (14 models)} & $\mathbf{87.7\% (0.023)}$ & \textbf{0.904 (0.026)}\\
		Two-step stacking (3 models) & $79.4\% (0.028)$ & 0.822 (0.030)\\
		Traditional stacking (14 models) & $81.8\% (0.033)$ & 0.854 (0.034)\\
		Traditional stacking (3 models) & $76.7\% (0.038)$ & 0.798 (0.037)\\
		Weighted voting (14 models) & $73.3\% (0.033)$ & 0.751 (0.040)\\
		Weighted voting (3 models) & $71.7\% (0.035)$ & 0.728 (0.037)\\
		Two-step stacking with GLPS only & $63.3\% (0.034)$ & 0.674 (0.047)\\
		\hline
	\end{tabular}}
\end{center}
\end{table}

\end{document}